%% file: main.tex
\documentclass{article}

\usepackage{subfigure}
\usepackage{multirow}
\usepackage[table,xcdraw]{xcolor} 
\usepackage{graphicx}

\usepackage[utf8]{inputenc} 
\usepackage[T1]{fontenc}    
\usepackage{hyperref}       
\usepackage{url}            
\usepackage{booktabs}       
\usepackage{amsfonts}       
\usepackage{nicefrac}       
\usepackage{microtype}      
\usepackage{xcolor}         
\usepackage{amsmath}    
\usepackage{amssymb}    
\usepackage{pifont}

\usepackage{hyperref}

\usepackage[accepted]{icml2025}

\usepackage{amsmath}
\usepackage{amssymb}
\usepackage{mathtools}
\usepackage{amsthm}

\usepackage[capitalize,noabbrev]{cleveref}

\theoremstyle{plain}

\theoremstyle{definition}

\theoremstyle{remark}

\usepackage[textsize=tiny]{todonotes}

\icmltitlerunning{Accurate Parameter-Efficient Test-Time Adaptation for Time Series Forecasting.}

\begin{document}

\twocolumn[
\icmltitle{Accurate Parameter-Efficient Test-Time Adaptation for Time Series Forecasting}


\icmlsetsymbol{equal}{*}

\begin{icmlauthorlist}
\icmlauthor{Heitor R. Medeiros}{equal,comp,sch}
\icmlauthor{Hossein Sharifi-Noghabi}{comp}
\icmlauthor{Gabriel L. Oliveira}{comp}
\icmlauthor{Saghar Irandoust}{comp}

\end{icmlauthorlist}

\icmlaffiliation{comp}{Borealis AI, Montreal, Canada}
\icmlaffiliation{sch}{Dept. of Systems Engineering, ETS Montreal, Canada}

\icmlcorrespondingauthor{Heitor R. Medeiros}{heitor.rapela-medeiros.1@ens.etsmtl.ca}

\icmlkeywords{Machine Learning, PUT ICML Workshop 2025}

\vskip 0.3in
]



\def\thefootnote{*}\footnotetext{Work done during an internship at Borealis AI.}

\printAffiliationsAndNotice{}

\input{sec/00_abstract}
\input{sec/01_intro}

\input{sec/02_related_works}

\input{sec/03_method}

\input{sec/04_experiments}

\input{sec/05_conclusion}

\bibliography{main.bib}
\bibliographystyle{icml2025}

\input{sec/06_appendix}

\end{document}

%% file: sec/00_abstract.tex
\begin{abstract}
Real-world time series often exhibit a non-stationary nature, degrading the performance of pre-trained forecasting models. Test-Time Adaptation (TTA) addresses this by adjusting models during inference, but existing methods typically update the full model, increasing memory and compute costs. We propose PETSA, a parameter-efficient method that adapts forecasters at test time by only updating small calibration modules on the input and output. PETSA uses low-rank adapters and dynamic gating to adjust representations without retraining. To maintain accuracy despite limited adaptation capacity, we introduce a specialized loss combining three components: (1) a robust term, (2) a frequency-domain term to preserve periodicity, and (3) a patch-wise structural term for structural alignment. PETSA improves the adaptability of various forecasting backbones while requiring fewer parameters than baselines. Experimental results on benchmark datasets show that PETSA achieves competitive or better performance across all horizons. Our code is available at: \url{https://github.com/BorealisAI/PETSA}.
\end{abstract}

%% file: sec/01_intro.tex
\section{Introduction}
\label{intro}

Time series forecasting (TSF) plays a critical role in applications such as weather prediction, traffic monitoring, and financial modeling~\citep{wu2021autoformer, zhou2021informer, kudrat2025patch}. While deep learning models like Transformers and MLPs have significantly improved TSF performance~\citep{shabani2022scaleformer, wang2025fredf}, they often assume stationarity and struggle when the data distribution shifts over time~\citep{kim2025battling}. In practice, such shifts, which are normally caused by seasonality, structural breaks, or domain shifts, lead to a significant degradation in accuracy~\citep{kim2024model, kim2025battling}.

Test-Time Adaptation (TTA) has emerged as a promising strategy to mitigate these shifts by updating models during inference~\citep{wang2020tent, kim2025battling}. However, most TTA methods either rely on access to source data~\citep{wang2020tent} or update the entire model~\citep{hu2022lora}, resulting in high computational overhead. Furthermore, limited information at test time makes reliable adaptation challenging~\citep{kim2024model, kudrat2025patch}

\begin{figure}[!t]
    \centering
    \includegraphics[width=0.8\linewidth]{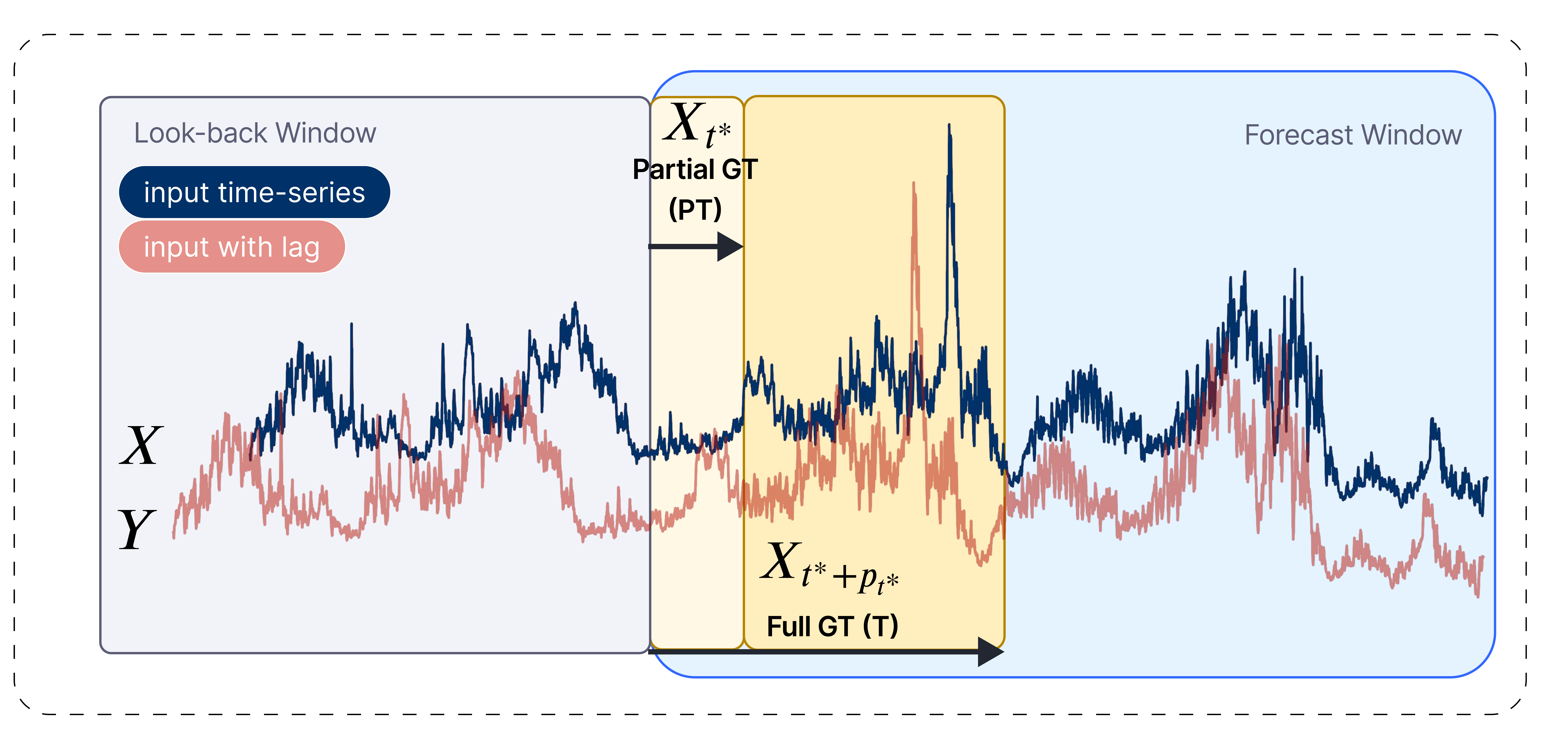}
    \vspace{-0.7em}
    \caption{\textbf{Illustration of the test-time adaptation setup in PETSA.} The model observes a look-back window and makes predictions over the forecast window. A partial portion of the ground truth (PT) becomes available shortly after prediction (light yellow), which is used to adapt the model online. Full ground truth (T) may also be observed after the forecast window completes (shaded yellow). PETSA uses both partial and delayed T to update lightweight calibration modules during inference. The $X$ is the time-series input, and $Y$ is the same time-series with a lag, which can be partially used as ground truth. The $X_{t^*}$ is the input at time ${t^*}$ and the partial batch goes until ${t^*}+{p_{t^*}}$ timestep. 
}
    \label{fig:forecast_window}
\end{figure}
\vspace{-0.7em}

\begin{figure*}[!ht]
    \centering
    \includegraphics[width=0.8\linewidth]{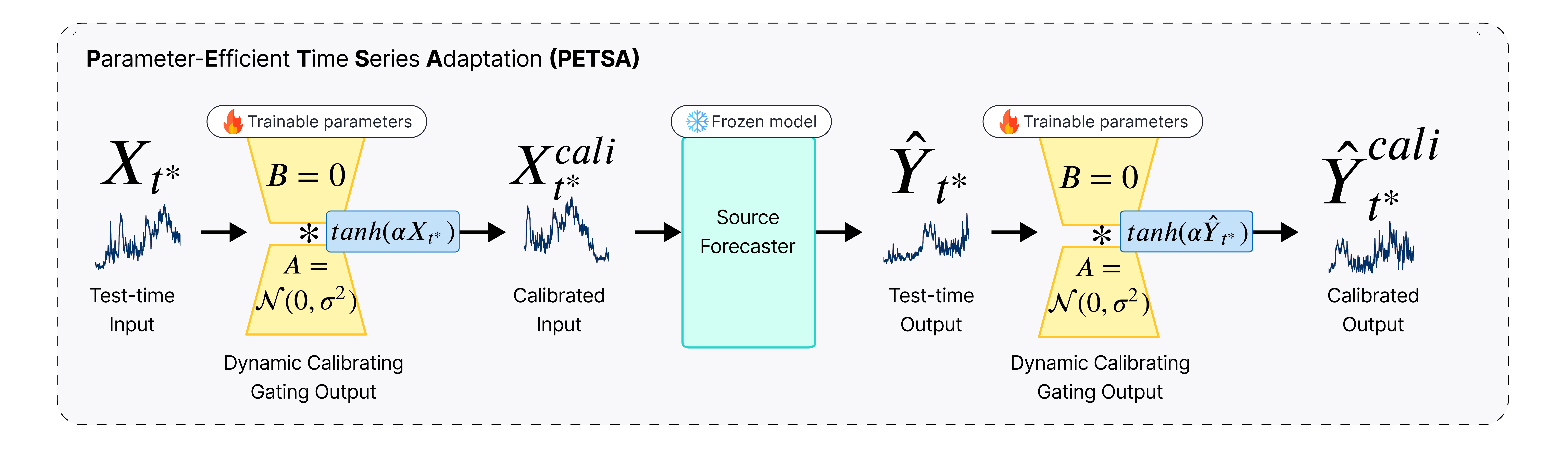}
    \vspace{-0.5em}
    \caption{\textbf{PETSA.} At test time, the input $X_{t^{}}$ is first passed through a dynamic input calibration module that applies a gated low-rank transformation. The calibrated input $X_{t^{}}^{\text{cali}}$ is then processed by a frozen pre-trained forecaster. Its output $\hat{Y}_{t^{*}}$ is refined by a similar output calibration module to produce the final prediction $\hat{Y}_{t^{*}}^{\text{cali}}$. Only the calibration modules are updated during test-time adaptation using the PETSA loss with partially and fully observed ground truth available with a delay. Modules with trainable parameters have a fire icon, while frozen ones have an ice icon.}
    \label{fig:petsa}
\end{figure*}

In this paper, we introduce Parameter‐Efficient Time-Series Adaptation (PETSA) framework (Figure~\ref{fig:petsa}), tailored for test-time adaptation of time-series forecasters. 

\noindent \textbf{Our main contributions can be summarized as follows.} 

\noindent \textbf{(1)} We propose PETSA, a test-time adaptation framework that calibrates input and output features using lightweight low-rank adapters and dynamic gating.

\noindent \textbf{(2)}  We design a unified PETSA loss combining Huber, frequency, and patch-wise structural terms for robust and structure-aware adaptation.

\noindent \textbf{(3)} We benchmark PETSA across six datasets and show that it improves multiple forecasters while maintaining high efficiency.

%% file: sec/02_related_works.tex
\section{Related Works}
\label{related_works}

\textbf{Time-Series Forecasting (TSF).} Recent TSF models span Transformers, linear projections, and MLP-based forecasters. Transformer-based models like iTransformer~\citep{liuitransformer} and PatchTST~\citep{nietime} capture long-range dependencies through self-attention, while linear approaches such as DLinear~\citep{zeng2023transformers} and OLS~\cite{toner2024analysis} offer competitive performance with lower complexity. MLP-based methods like FreTS~\citep{yi2023frequency} and MICN~\citep{wang2023micn} balance expressiveness and efficiency using global/local mixing. These models highlight the trade-off between accuracy and computational cost in TSF.

\textbf{Parameter-Efficient Fine-Tuning (PEFT).} PEFT techniques adapt large models using a small number of tunable parameters. Popular strategies include LoRA~\citep{hu2022lora}, DoRA~\citep{liu2024dora}, and visual adapters like VPT~\citep{jia2022visual} or AdaptFormer~\citep{chen2022adaptformer}. While PEFT has seen wide use in vision and NLP, recent efforts extend to TSF~\citep{gupta2024low, ruan2024low, nie2024channel}. However, existing methods mainly focus on fine-tuning and do not address test-time adaptation.

\textbf{Test-Time Adaptation (TTA).} TTA enables models to adapt to distribution shifts during inference using unlabeled data~\citep{zhao2023pitfalls, liang2025comprehensive}. Techniques like TENT~\citep{wang2020tent}, LAME~\citep{boudiaf2022parameter}, and entropy minimization update model statistics or outputs. In TSF, TAFAS~\citep{kim2025battling} introduces a batch-level adaptation scheme using delayed partial labels. PETSA builds on this line by introducing a parameter-efficient, gating-based architecture with specialized losses for robust and structured test-time adaptation.

\newcommand{\cmark}{\checkmark}
\newcommand{\xmark}{\ding{55}}
\newcolumntype{g}{>{\columncolor{yellow!15}}c}
\newcolumntype{q}{>{\columncolor{yellow!15}}l}

\begin{table*}[t]
  \caption{\textbf{MSE across datasets and window sizes}. The input training sequence length is set to $96$ for all baselines. Results for \xmark - checkpoint, TF - TAFAS, and PETSA - PT. The lower MSE is marked in bold. Additionally, we provided a row-counter (RW), which counts the winner for each row, meaning the best for the window length on the dataset among all models, and a column-counter (CW), with the winner per model, and the total sum of column winners.}

  \label{tab:main-table}
  \centering
  \resizebox{0.95\textwidth}{!}{%

  \begin{tabular}{lclllllllllllllllllllll} 
    \toprule
    \rowcolor{white}
    {} & {} & \multicolumn{6}{c}{Transformer-based} 
    & \multicolumn{6}{c}{Linear-based} & \multicolumn{6}{c}{MLP-based} & \multicolumn{2}{c}{Counter} \\
    \cmidrule(r){3-8}
    \cmidrule(r){9-14}
    \cmidrule(r){15-20}
    \rowcolor{white}
    \multicolumn{2}{c}{Models} & \multicolumn{3}{c}{iTransformer}
    & \multicolumn{3}{c}{PatchTST} & \multicolumn{3}{c}{DLinear} & \multicolumn{3}{c}{OLS} & \multicolumn{3}{c}{FreTS} & \multicolumn{3}{c}{MICN} & \multicolumn{2}{c}{RW} \\
    
    \cmidrule(r){3-5}
    \cmidrule(r){6-8}
    \cmidrule(r){9-11}
    \cmidrule(r){12-14}
    \cmidrule(r){15-17}
    \cmidrule(r){18-20}
    \cmidrule(r){21-22}
    \rowcolor{white}
    {} & Wind. & \xmark & TF & PT &  \xmark & TF & PT &  \xmark & TF & PT &  \xmark & TF & PT &  \xmark & TF & PT &  \xmark & TF & PT & TF & PT \\
    
    \toprule

    {} &	96 & 0.449 & 0.435 & \textbf{0.432} & 0.433 & \textbf{0.426} & \textbf{0.426} & 0.470 & 0.462 & \textbf{0.459} & 0.451 & 0.442 & \textbf{0.440} & 0.446 & 0.440 & \textbf{0.438} & 0.520 & \textbf{0.493} & \textbf{0.493} & 2 & \textbf{6} \\

    {} &	192 & 0.510 & 0.503 & \textbf{0.501} & 0.491 & 0.482 & \textbf{0.481} & 0.521 & 0.512 & \textbf{0.511} & 0.505 & \textbf{0.492} & \textbf{0.492} & 0.502 & 0.494 & \textbf{0.492} & 0.591 & 0.560 & \textbf{0.559} & 1 & \textbf{6} \\

    \multirow{4}{*}[1.3em]{\rotatebox{90}{ETTh1}} &	336 & 0.564 & 0.562 & \textbf{0.561} & 0.555 & 0.546 & \textbf{0.543} & 0.566 & 0.560 & \textbf{0.555} & 0.551 & 0.542 & \textbf{0.538} & 0.554 & 0.548 & \textbf{0.547} & 0.665 & \textbf{0.632} & 0.643 & 1 & \textbf{5} \\

    {} &	720 & 0.702 & 0.663 & \textbf{0.659} & 0.706 & \textbf{0.680} & \textbf{0.680} & 0.712 & 0.682 & \textbf{0.679} & 0.700 & 0.666 & \textbf{0.650} & 0.718 & \textbf{0.687} & 0.688 & 0.904 & 0.792 & \textbf{0.785} & 2 & \textbf{5} \\
    \cmidrule(r){2-22}\rowcolor{yellow!10}\cellcolor{white}	& Avg &	0.557 & 0.541 & \textbf{0.538} & 0.546 & 0.533 & \textbf{0.532} & 0.567 & 0.554 & \textbf{0.551} & 0.552 & 0.535 & \textbf{0.530} & 0.555 & 0.542 & \textbf{0.541} & 0.670 & \textbf{0.619} & 0.620 & 1 & \textbf{5} \\

    \midrule																			
    {} &	96 & 0.439 & 0.416 & \textbf{0.413} & 0.451 & 0.437 & \textbf{0.436} & 0.444 & 0.417 & \textbf{0.414} & 0.444 & 0.416 & \textbf{0.415} & 0.433 & 0.421 & \textbf{0.416} & 0.487 & 0.458 & \textbf{0.456} & 0 & \textbf{6} \\

    {} &	192 & 0.508 & 0.476 & \textbf{0.473} & 0.504 & \textbf{0.486} & 0.489 & 0.518 & 0.480 & \textbf{0.474} & 0.518 & 0.479 & \textbf{0.475} & 0.501 & 0.482 & \textbf{0.475} & 0.554 & 0.511 & \textbf{0.510} & 1 & \textbf{5} \\

    \multirow{4}{*}[1.3em]{\rotatebox{90}{ETTm1}} &	336 & 0.613 & 0.556 & \textbf{0.552} & 0.558 & \textbf{0.539} & 0.542 & 0.593 & 0.549 & \textbf{0.545} & 0.593 & 0.548 & \textbf{0.543} & 0.570 & 0.547 & \textbf{0.543} & 0.612 & 0.579 & \textbf{0.573} & 1 & \textbf{5} \\

    {} &	720 & 0.485 & 0.453 & \textbf{0.450} & 0.479 & \textbf{0.463} & 0.465 & 0.482 & 0.449 & \textbf{0.446} & 0.481 & 0.449 & \textbf{0.446} & 0.468 & 0.452 & \textbf{0.448} & 0.525 & 0.486 & \textbf{0.484} & 1 & \textbf{5} \\

    \cmidrule(r){2-22}\rowcolor{yellow!10}\cellcolor{white}	& Avg & 0.257 & 0.255 & \textbf{0.254} & 0.236 & \textbf{0.235} & \textbf{0.235} & 0.232 & \textbf{0.230} & \textbf{0.230} & 0.231 & \textbf{0.228} & \textbf{0.228} & 0.239 & \textbf{0.236} & \textbf{0.236} & 0.256 & \textbf{0.252} & \textbf{0.252} & 5 & \textbf{6} \\

    \midrule																			

    {} &	96 & 0.344 & 0.330 & \textbf{0.328} & 0.317 & \textbf{0.308} & 0.309 & 0.325 & 0.319 & \textbf{0.318} & 0.326 & 0.319 & \textbf{0.318} & 0.332 & \textbf{0.321} & \textbf{0.321} & 0.359 & \textbf{0.339} & 0.342 & 3 & \textbf{4} \\
    
    {} &	192 & 0.424 & \textbf{0.396} & 0.397 & 0.433 & \textbf{0.402} & \textbf{0.402} & 0.409 & 0.387 & \textbf{0.385} & 0.416 & 0.391 & \textbf{0.388} & 0.412 & \textbf{0.383} & \textbf{0.383} & 0.437 & 0.439 & \textbf{0.434} & 3 & \textbf{5} \\
    
    \multirow{4}{*}[1.3em]{\rotatebox{90}{ETTh2}} &	336 & 0.332 & 0.320 & \textbf{0.319} & 0.318 & \textbf{0.305} & \textbf{0.305} & 0.313 & \textbf{0.305} & \textbf{0.305} & 0.314 & 0.305 & \textbf{0.304} & 0.317 & \textbf{0.306} & \textbf{0.306} & 0.345 & \textbf{0.334} & 0.335 & 4 & \textbf{5} \\

    {} &	720 & 0.168 & 0.167 & \textbf{0.166} & \textbf{0.160} & \textbf{0.160} & \textbf{0.160} & 0.160 & \textbf{0.158} & \textbf{0.158} & 0.160 & \textbf{0.159} & \textbf{0.159} & 0.158 & \textbf{0.157} & \textbf{0.157} & \textbf{0.175} & \textbf{0.175} & 0.176 & \textbf{5} & \textbf{5} \\

    \cmidrule(r){2-22}\rowcolor{yellow!10}\cellcolor{white}	& Avg & 0.220 & 0.217 & \textbf{0.215} & 0.207 & \textbf{0.204} & \textbf{0.204} & 0.193 & \textbf{0.191} & \textbf{0.191} & 0.194 & \textbf{0.192} & \textbf{0.192} & 0.192 & \textbf{0.191} & \textbf{0.191} & 0.213 & 0.209 & \textbf{0.203} & 4 & \textbf{6} \\

    \midrule

    {} &	96 &	0.339 & 0.330 & \textbf{0.322} & 0.334 & \textbf{0.327} & 0.328 & 0.306 & 0.297 & \textbf{0.296} & 0.307 & \textbf{0.298} & \textbf{0.298} & 0.301 & \textbf{0.292} & 0.293 & 0.332 & 0.322 & \textbf{0.320} & 3 & \textbf{4} \\

    {} &	192 & 0.250 & 0.244 & \textbf{0.241} & 0.237 & \textbf{0.235} & \textbf{0.235} & 0.223 & \textbf{0.219} & \textbf{0.219} & 0.223 & \textbf{0.220} & \textbf{0.220} & 0.221 & \textbf{0.217} & 0.218 & 0.243 & 0.238 & \textbf{0.236} & 4 & \textbf{5} \\

    \multirow{4}{*}[1.3em]{\rotatebox{90}{ETTm2}} &	336 &	0.087 & \textbf{0.085} & 0.086 & 0.086 & \textbf{0.082} & 0.083 & 0.091 & 0.089 & \textbf{0.088} & 0.081 & 0.080 & \textbf{0.078} & 0.083 & \textbf{0.079} & \textbf{0.079} & 0.115 & 0.115 & \textbf{0.109} & 3 & \textbf{4} \\

    {} &	720 &	0.181 & \textbf{0.174} & 0.175 & 0.188 & \textbf{0.174} & 0.179 & 0.183 & 0.176 & \textbf{0.173} & 0.173 & \textbf{0.164} & 0.165 & 0.173 & 0.164 & \textbf{0.163} & 0.216 & \textbf{0.198} & \textbf{0.198} & \textbf{4} & 3 \\

    \cmidrule(r){2-22}\rowcolor{yellow!10}\cellcolor{white}	& Avg &	0.343 & \textbf{0.313} & 0.335 & 0.338 & \textbf{0.281} & 0.332 & 0.328 & 0.294 & \textbf{0.292} & 0.323 & 0.285 & \textbf{0.281} & 0.324 & \textbf{0.295} & 0.298 & 0.398 & 0.304 & \textbf{0.280} & \textbf{3} & \textbf{3}  \\
    
    \midrule																			

    {} &	96 &	0.366 & 0.345 & \textbf{0.341} & 0.372 & \textbf{0.353} & 0.367 & 0.372 & 0.359 & \textbf{0.357} & 0.353 & 0.294 & \textbf{0.286} & 0.354 & 0.335 & \textbf{0.327} & 0.558 & \textbf{0.307} & 0.357 & 2 & \textbf{4} \\

    {} &	192 & 0.173 & \textbf{0.166} & \textbf{0.166} & 0.173 & \textbf{0.170} & 0.171 & 0.195 & 0.180 & \textbf{0.176} & 0.196 & 0.181 & \textbf{0.178} & 0.186 & 0.175 & \textbf{0.174} & 0.176 & 0.175 & \textbf{0.174} & 2 & \textbf{5} \\

    \multirow{4}{*}[1.3em]{\rotatebox{90}{Exchange}} &	336 & 0.223 & \textbf{0.211} & 0.212 & 0.220 & \textbf{0.214} & 0.216 & 0.240 & 0.224 & \textbf{0.223} & 0.241 & \textbf{0.222} & 0.223 & 0.231 & \textbf{0.215} & 0.218 & 0.224 & \textbf{0.217} & 0.220 & \textbf{5} & 1 \\

    {} &	720 &	0.281 & \textbf{0.261} & 0.265 & 0.276 & \textbf{0.265} & 0.268 & 0.292 & \textbf{0.271} & \textbf{0.271} & 0.292 & \textbf{0.271} & 0.273 & 0.284 & \textbf{0.264} & 0.266 & 0.281 & 0.269 & \textbf{0.268} & \textbf{5} & 2 \\

    \cmidrule(r){2-22}\rowcolor{yellow!10}\cellcolor{white}	& Avg &	0.355 & \textbf{0.339} & 0.341 & 0.355 & 0.337 & \textbf{0.336} & 0.364 & 0.350 & \textbf{0.345} & 0.364 & \textbf{0.344} & 0.346 & 0.360 & \textbf{0.340} & 0.344 & 0.353 & 0.347 & \textbf{0.345} & \textbf{3} & \textbf{3} \\

    \midrule									
    {} &	96 & 0.173 & \textbf{0.166} & \textbf{0.166} & 0.173 & \textbf{0.170} & 0.171 & 0.195 & 0.180 & \textbf{0.176} & 0.196 & 0.181 & \textbf{0.178} & 0.186 & 0.175 & \textbf{0.174} & 0.176 & 0.175 & \textbf{0.174} & 2 & \textbf{5}  \\

    {} &	192 &	0.223 & \textbf{0.211} & 0.212 & 0.220 & \textbf{0.214} & 0.216 & 0.240 & 0.224 & \textbf{0.223} & 0.241 & \textbf{0.222} & 0.223 & 0.231 & \textbf{0.215} & 0.218 & 0.224 & \textbf{0.217} & 0.220 & \textbf{5} & 1 \\

    \multirow{4}{*}[1.3em]{\rotatebox{90}{Weather}} &	336 &	0.281 & \textbf{0.261} & 0.265 & 0.276 & \textbf{0.265} & 0.268 & 0.292 & \textbf{0.271} & \textbf{0.271} & 0.292 & \textbf{0.271} & 0.273 & 0.284 & \textbf{0.264} & 0.266 & 0.281 & 0.269 & \textbf{0.268} & \textbf{5} & 2 \\

    {} &	720 &	0.355 & \textbf{0.339} & 0.341 & 0.355 & 0.337 & \textbf{0.336} & 0.364 & 0.350 & \textbf{0.345} & 0.364 & \textbf{0.344} & 0.346 & 0.360 & \textbf{0.340} & 0.344 & 0.353 & 0.347 & \textbf{0.345}  & \textbf{3} & \textbf{3} \\

    \cmidrule(r){2-22}\rowcolor{yellow!10}\cellcolor{white}	& Avg &	0.258 & \textbf{0.244} & 0.246 & 0.256 & \textbf{0.247} & 0.248 & 0.273 & 0.256 & \textbf{0.254} & 0.273 & \textbf{0.255} & \textbf{0.255} & 0.265 & \textbf{0.248} & 0.251 & 0.258 & \textbf{0.252} & \textbf{0.252} & \textbf{5} & 3 \\

    \midrule
    
    \rowcolor{white}
    \multicolumn{2}{c}{} & \xmark & TF & PT &  \xmark & TF & PT &  \xmark & TF & PT &  \xmark & TF & PT &  \xmark & TF & PT &  \xmark & TF & PT \\
    
     \cmidrule(r){2-20}

    Counter & CW & {} & 13 & \textbf{19} & {} & \textbf{24} & 14 & {} & 7 & \textbf{30} & {} & 14 & \textbf{23} & {} & 18 & \textbf{19} & {} & 12 & \textbf{22} \\

     \cmidrule(r){1-20}
    \textbf{Sum Col.} & TF: 88 & PT: \textbf{127} \\

    \bottomrule
  \end{tabular}
  }
\end{table*}

%% file: sec/03_method.tex
\section{Proposed Method}
\label{method}

\subsection{Preliminary Definitions}

\textbf{TSF.} TSF involves predicting future values of a sequence based on historical observations. Formally, given a historical multivariate time series \( X = \{ x_{t-L}, x_{t-L+1}, \dots, x_{t-1} \} \) consisting of \( L \) consecutive observations, the goal of TSF is to learn a forecasting model \( f_\theta(\cdot) \) that generates accurate predictions of the next \( T \) future steps, denoted as: $Y = \{ x_{t}, x_{t+1}, \dots, x_{t+T-1} \} =  f_\theta(X)$.

\textbf{TTA in TSF.} In TSF, TTA mitigates distribution shifts by updating the model using only test inputs. Methods like TAFAS assume that partial ground truth becomes available shortly after prediction, enabling online updates. The adaptation window is defined using the dominant period, estimated via Fast Fourier Transform (FFT). PETSA adopts this setup, using both partial and full labels to update its lightweight gating modules during inference, same as in TAFAS~\citep{kim2025battling}.

\subsection{PETSA}

We propose Parameter-Efficient Time-Series Adaptation (PETSA), a lightweight framework, designed to adapt time-series forecasting models at inference without modifying the core model parameters. It introduces input and output calibration modules that leverage low-rank adapters and dynamic gating mechanisms to correct for distribution shifts. \textbf{Dynamic Calibration Mechanism.} At test time, PETSA calibrates both the input and output of a frozen forecaster using lightweight low-rank adapters and dynamic gating, inspired by Dynamic Tanh (DyT)~\citep{zhu2025transformers} and TAFAS. The calibrated input ($\hat{X}_{t^{*}}^{\text{cali}}$) and calibrated output ($\hat{Y}_{t^{*}}^{\text{cali}}$) are computed as follows:

\begin{equation}
\begin{aligned}
    \hat{X}_{t^{*}}^{\text{cali}} &= X_{t^{*}} + \left( \tanh(\alpha \odot X_{t^{*}}) \cdot W + b \right) \\
    \hat{Y}_{t^{*}}^{\text{cali}} &= \hat{Y}_{t^{*}} + \left( \tanh(\alpha \odot \hat{Y}_{t^{*}}) \cdot W + b \right),
\end{aligned}
\label{eq:dynamic_calibration}
\end{equation}

where ${X}_{t^{*}} \in \mathbb{R}^{B \times L \times V}$ and \(\hat{Y}_{t^{*}} \in \mathbb{R}^{B \times L \times V}\) are the test-time input and output respectively, $\alpha \in \mathbb{R}^{V}$ is a learnable gating parameter per variable (we control the initialization with a hyperparameter), applied element-wise, $W = A \cdot B$, with $A \in \mathbb{R}^{L \times r}$, \(B \in \mathbb{R}^{r \times L \times V}\) forming the low-rank weight tensor, $b \in \mathbb{R}^{L \times V}$is a learnable bias term ($A$ is initialized with Xavier Norm. and $B$ with zeros). This enables PETSA to efficiently calibrate time-series representations by updating only a small number of parameters at test time. 

\textbf{PETSA Optimization.} PETSA uses a combination of different losses, while TAFAS only uses MSE loss. Our PETSA loss combines total and partial losses ($\mathcal{L}_{PETSA} = \mathcal{L}_{T} + \mathcal{L}_{pt}$) , where \(\mathcal{L}_T\) is computed using delayed full ground-truth labels and \(\mathcal{L}_{pt}\) uses partially observed labels~\citep{kim2025battling}. Each loss term incorporates three components: (1) a Huber loss (\(\mathcal{L}_{\text{Hub}}\)) ~\citep{huber1992robust} for robustness to outliers~\citep{shabani2022scaleformer}, described as follows:

\begin{align}
\mathcal{L}_{Hub} =
\begin{cases}
0.5 (\hat{Y}_{t^{*}}^{cali} - Y_{t^{*}})^2, \text{if } |\hat{Y}_{t^{*}}^{cali} - Y_{t^{*}}| < \delta \\
\delta \cdot (|\hat{Y}_{t^{*}}^{cali} - Y_{t^{*}}| - 0.5 \cdot \delta), \text{otherwise}.
\end{cases} \label{eq:hub_loss}
\end{align}

where $\delta$ is a hyperparameter to control the sensitivity to outliers and smoothness of the predictions (in this work, $\delta$ is fixed at $0.5$), (2) a frequency-domain loss (\(\mathcal{L}_{\text{freq}}\)) that aligns the FFT spectra of predictions and ground truth to preserve periodic patterns, while reducing estimation bias, as described in FreDF~\citep{wang2025fredf}, described as follows:

\vspace{-0.7em}
\begin{align}
\mathcal{L}_{freq}  = \left\lVert \mathcal{F}(\hat{Y}_{t^{*}}^{cali}) - \mathcal{F}(Y_{t^{*}}) \right\rVert_1, \label{eq:freq_loss}
\end{align}
\vspace{-0.7em}

where $\mathcal{F}(.) = FFT$, and (3) a patch-wise structural loss (\(\mathcal{L}_{\text{pw}}\)) that captures local correlations, means, and variances to enhance structural alignment~\citep{kudrat2025patch}, described as follows:

\vspace{-0.5em}
\begin{equation}
\mathcal{L}_{pw} = \sum_{k \in \{\text{corr}, \text{mean}, \text{var}\}} \mathcal{L}_k(\hat{Y}^{cali}_{t^*}, Y_{t^*}).
\label{eq:patch_loss}
\end{equation}
\vspace{-0.5em}

Finally, the partial (PT) and delayed GT (T) loss are described as follows:
 \vspace{-0.5em}
 \begin{align}
    \mathcal{L}_{pt} &= \mathcal{L}_{Hub_{pt}} + \mathcal{L}_{pw_{pt}} + \beta \mathcal{L}_{freq_{pt}} \label{eq:pt_loss} \\
    \mathcal{L}_{T} &= \mathcal{L}_{Hub_{T}} + \mathcal{L}_{pw_{T}} + \beta \mathcal{L}_{freq_{T}}.
    \label{eq:T_loss}
\end{align}
 
To the best of our knowledge, this is the first work to do parameter-efficient TTA for TSF. By updating only a small set of calibration parameters at test time, PETSA enables fast, stable, and memory-efficient adaptation across a wide range of forecasting models and datasets.

%% file: sec/04_experiments.tex
\begin{figure}[!ht]
    \centering
    \includegraphics[width=0.75\linewidth]{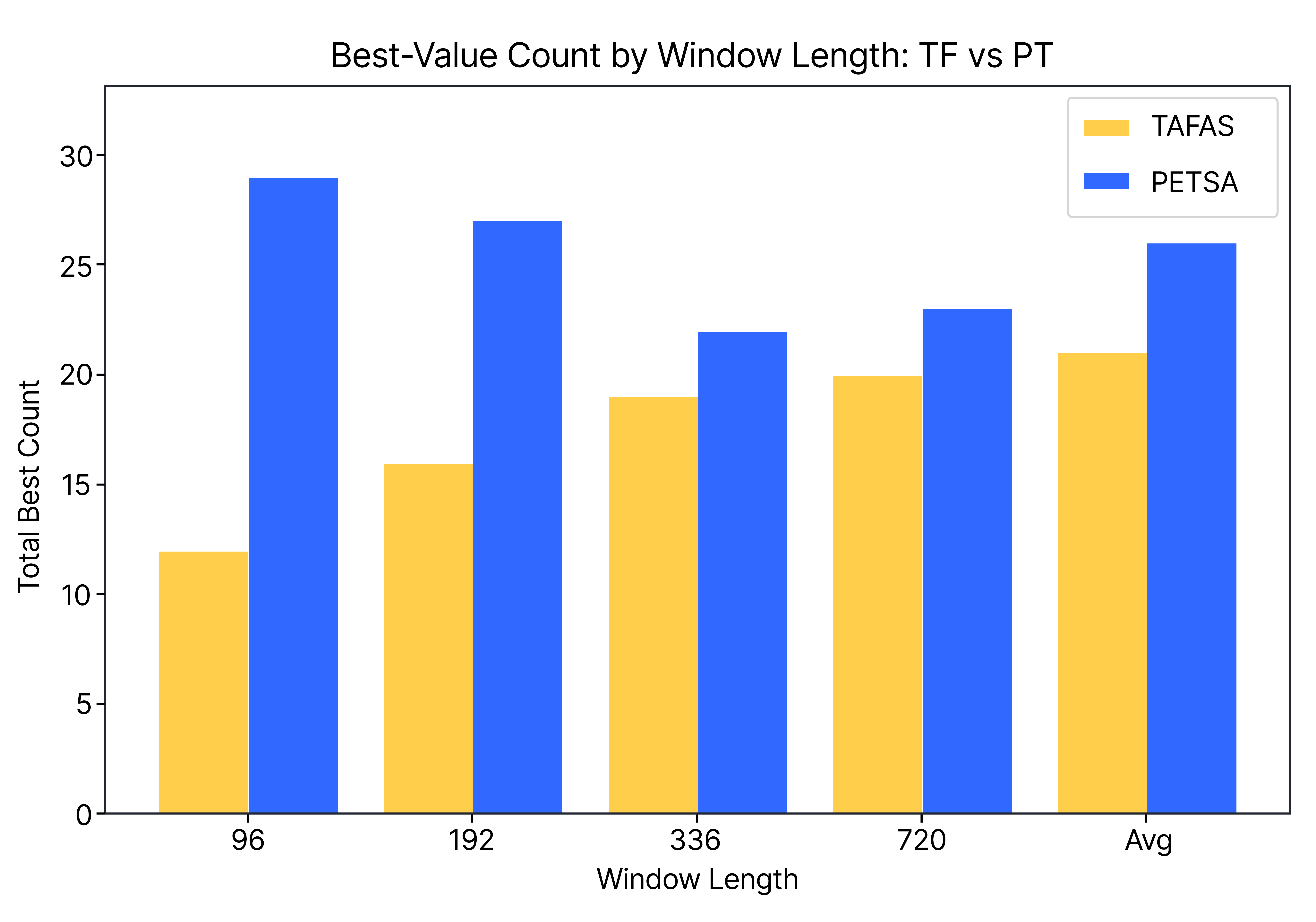}
    \vspace{-0.5em}
    \caption{\textbf{Total number of best-value wins grouped by window length for TAFAS and PETSA approaches.} PETSA consistently outperforms TAFAS across all horizons.}
    \label{fig:tf-pt-by-window}
\end{figure}
\vspace{-0.5em}

\section{Experiments}
\label{experiments}

\subsection{Experimental Protocol}

\textbf{(a) Datasets:} We demonstrate the effectiveness of our method, PETSA, using widely used multivariate TSF benchmark datasets: ETTh1, ETTm1, ETTh2, ETTm2, Exchange, and Weather~\citep{wu2021autoformer, zhou2021informer}.

\textbf{(b) Implementation Details:} Our framework is built on top of TAFAS~\citep{kim2025battling}. We used PyTorch for PETSA implementation, and training/adapt the models using one NVIDIA A100. 

\textbf{(c) Baselines:}  We evaluate our proposed method against a diverse set of baseline models, grouped into three main categories: (1) Transformer-based approaches, including iTransformer~\citep{liuitransformer}, PatchTST~\citep{nietime}; (2) Linear-based models, comprising DLinear~\citep{zeng2023transformers}, OLS~\citep{toner2024analysis}, and (3) MLP-based influential architectures, such as FreTS~\citep{yi2023frequency}, MICN~\citep{wang2023micn}. Additionally, we provide the methods without and with adaptation using TAFAS~\citep{kim2025battling} and  PETSA. We provide additional details in the Appendix.

\begin{figure}[!ht]
    \centering
    \includegraphics[width=0.75\linewidth]{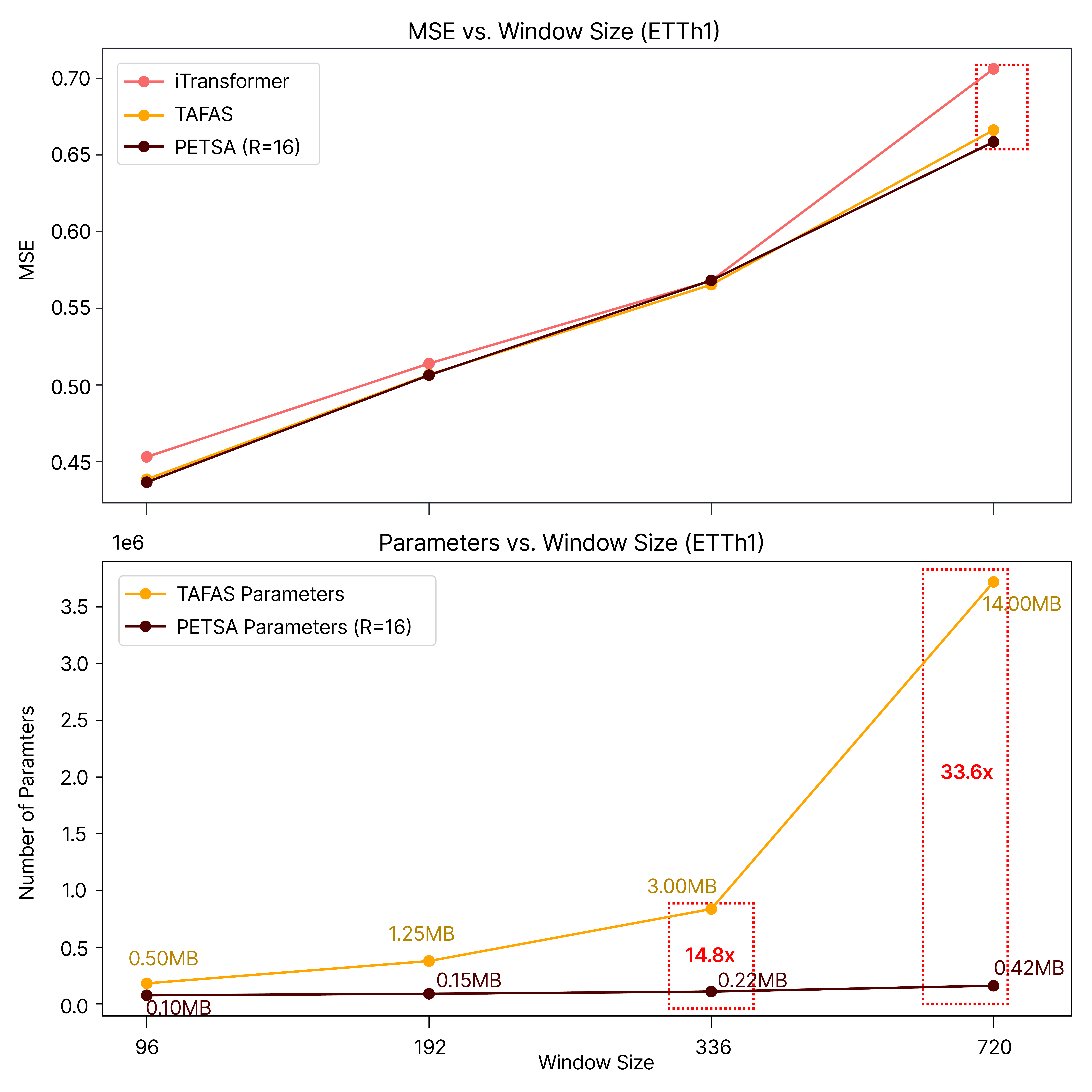}
    \vspace{-0.5em}
    \caption{\textbf{Comparison of PETSA and TAFAS on ETTh1 for iTransformer.} Top: MSE across different window sizes with no adaptation, TAFAS, and PETSA. Bottom: Number of trainable parameters used for adaptation. PETSA achieves similar or better accuracy while using up to \textbf{33.6$\times$} fewer parameters at window size 720. Memory usage is annotated in MB.}
    \label{fig:petsa-tafas-comparison}
\end{figure}
\vspace{-0.5em}

\subsection{Results}

In Table~\ref{tab:main-table}, across all datasets and model categories, PETSA achieves the highest number of best-MSE scores ($127$ wins), outperforming TAFAS ($88$ wins). Its consistent advantage across transformer-, linear-, and MLP-based architectures demonstrates strong adaptability, where all PETSA models had fewer parameters than TAFAS. Figure~\ref{fig:tf-pt-by-window} shows that PETSA achieves more best-value scores than TAFAS across different window lengths. Even as the forecast window increases, PETSA maintains a strong lead, demonstrating robustness to longer-term uncertainty. In Figure~\ref{fig:petsa-tafas-comparison}, PETSA achieves consistently lower MSE across all window sizes, and for window size 720, it has 33× fewer parameters than TAFAS, highlighting its efficiency due to the low-rank adaptation with dynamic gating, which is input conditioned and more robust to outliers in long-range forecasting as a result of its loss optimization.

%% file: sec/05_conclusion.tex
\section{Conclusion}
\label{conclusion}

In this work, we introduced PETSA, a lightweight, parameter‐efficient test‐time adaptation framework for time‐series forecasting that dynamically corrects both inputs and outputs via gated calibration modules. PETSA test-time calibration loss combines a robust component, a frequency‐domain term to preserve dominant periodic patterns, and a patch‐wise structural term to enforce structural alignment, which are essential to adapt the forecaster during test-time. Across diverse benchmarks, PETSA consistently improves forecasting performance while updating fewer parameters against baselines.

%% file: sec/06_appendix.tex
\newpage

\section*{Appendix}

Here we provide additional information and results for our paper.

\textbf{Reproducing the results}

Our codebase is built on top of TAFAS~\citep{kim2025battling}, where we followed their experimental setup and hyperparameters for generating the baseline checkpoint models and adapted models. Additionally, for our method, we have hyperparameters to control the frequency loss, the number of low-rank parameters, and the gating initialization, where we provide additional ablations in the next session.

\textbf{Additional results on Parameter-Efficiency}

\begin{figure*}[!ht]
    \centering
    \includegraphics[width=1.0\linewidth]{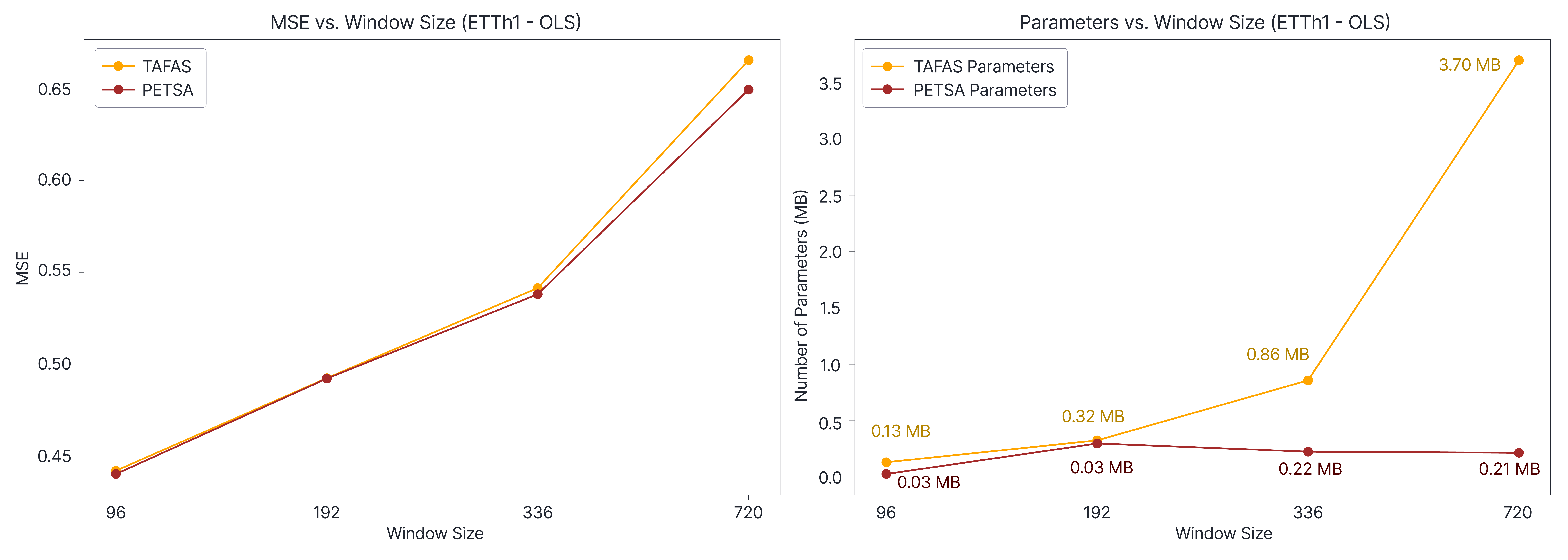}
    
    \caption{\textbf{Comparison of PETSA and TAFAS on ETTh1 for OLS.} Left: MSE across different window sizes with TAFAS, and PETSA. Right: Number of trainable parameters used for adaptation. Memory usage is annotated in MB.}
    \label{fig:param_ETTh1_OLS}
\end{figure*}

In Figure~\ref{fig:param_ETTh1_OLS}, PETSA and TAFAS show very similar MSE results across all window sizes on the ETTh1 dataset using the OLS model. Both methods follow the same trend, with PETSA slightly outperforming TAFAS at larger windows. In terms of parameters, PETSA remains highly efficient, using only $0.21$ MB at window size 720, while TAFAS requires $3.70$ MB. Across all window sizes, PETSA keeps memory usage consistently low while achieving comparable or better performance, highlighting its parameter efficiency.

\begin{figure*}[!ht]
    \centering
    \includegraphics[width=1.0\linewidth]{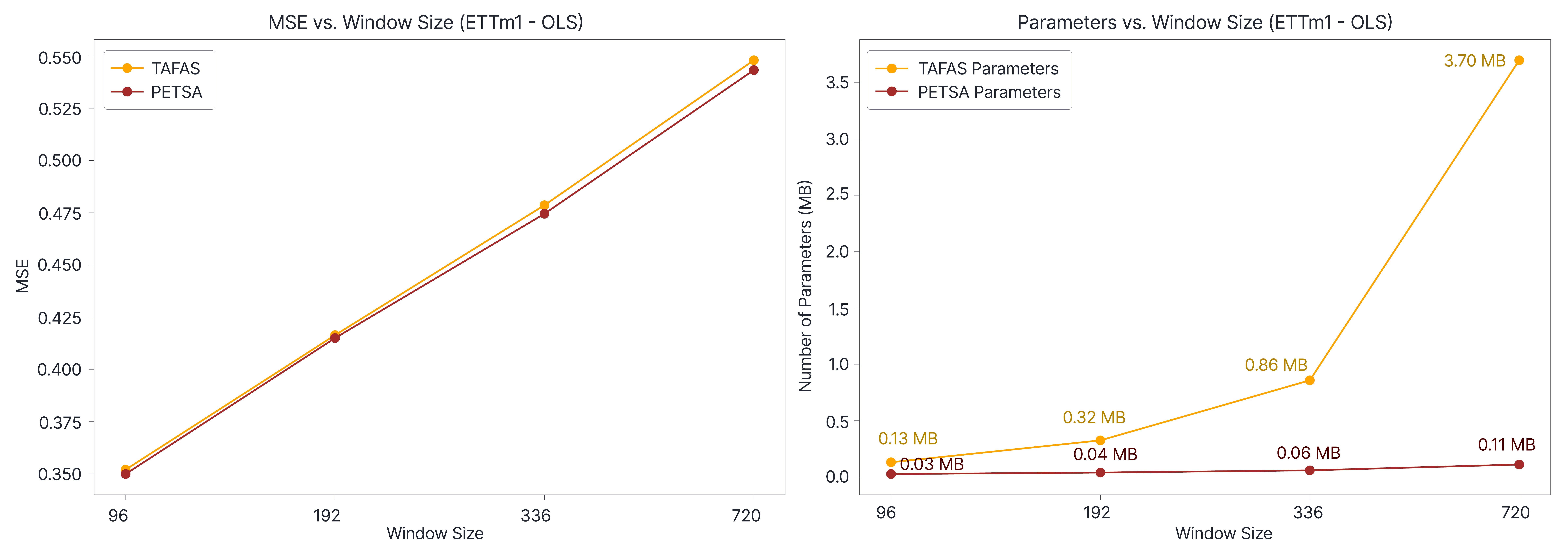}
    
    \caption{\textbf{Comparison of PETSA and TAFAS on ETTm1 for OLS.} Left: MSE across different window sizes with TAFAS, and PETSA. Right: Number of trainable parameters used for adaptation. Memory usage is annotated in MB.}
    \label{fig:param_ETTm1_OLS}
\end{figure*}

In Figure~\ref{fig:param_ETTm1_OLS}, PETSA achieves similar MSE to TAFAS across all window sizes on ETTm1 with OLS, with slightly better results at short horizons. For a window of 720, we kept the memory low at $0.11$MB, while TAFAS required $3.70$. As this dataset is easier than the other one, we had a good trade-off between performance and memory.

\begin{figure*}[!ht]
    \centering
    \includegraphics[width=1.0\linewidth]{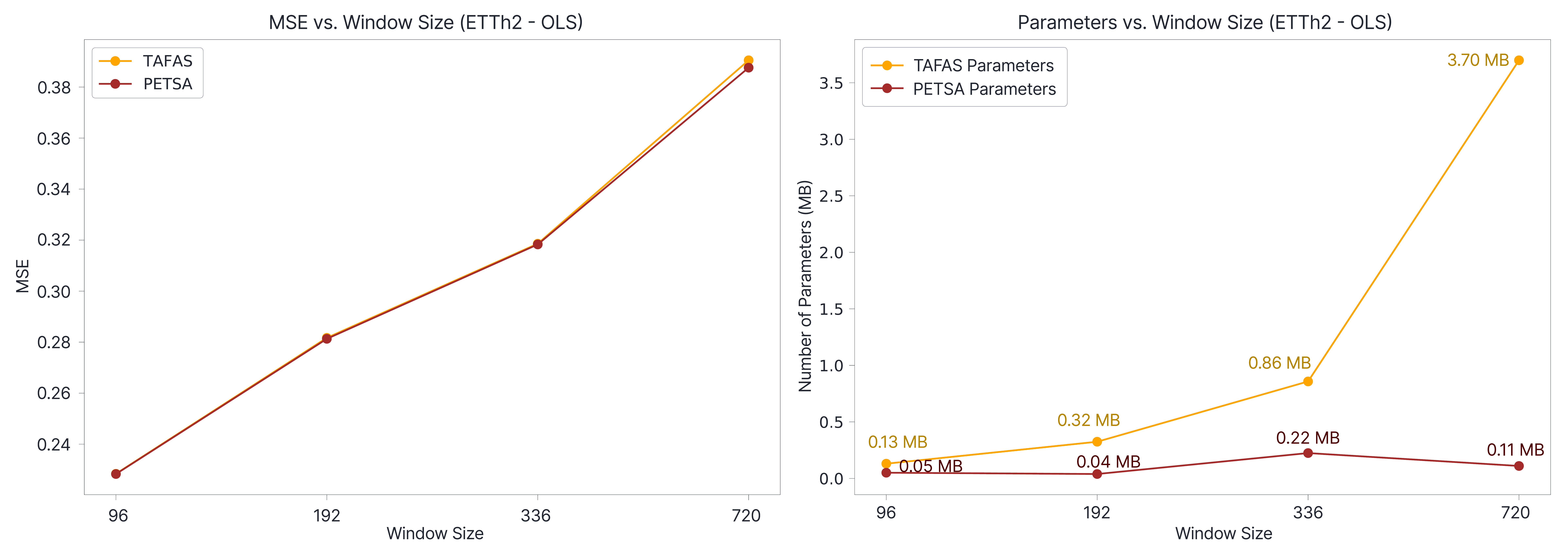}
    
    \caption{\textbf{Comparison of PETSA and TAFAS on ETTh2 for OLS.} Left: MSE across different window sizes with TAFAS, and PETSA. Right: Number of trainable parameters used for adaptation. Memory usage is annotated in MB.}
    \label{fig:param_ETTh2_OLS}
\end{figure*}

\begin{figure*}[!ht]
    \centering
    \includegraphics[width=1.0\linewidth]{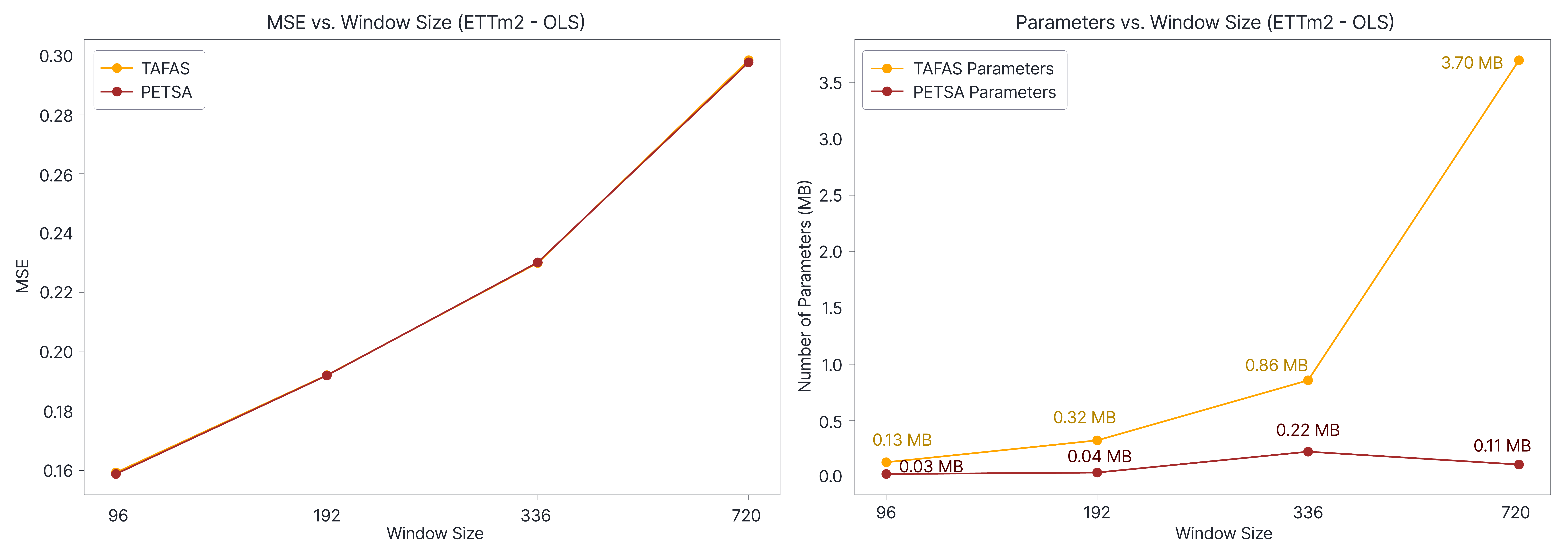}
    
    \caption{\textbf{Comparison of PETSA and TAFAS on ETTm2 for OLS.} Left: MSE across different window sizes with TAFAS, and PETSA. Right: Number of trainable parameters used for adaptation. Memory usage is annotated in MB.}
    \label{fig:param_ETTm2_OLS}
\end{figure*}

We had similar trends for Figures~\ref{fig:param_ETTh2_OLS} and Figures~\ref{fig:param_ETTm2_OLS}, we are comparable with TAFAS in terms of MSE, and the memory is also lower. However, we can see that ETTh1/ETTh2 requires a bit more memory than ETTm1/ETTm2 to achieve competitive results. This trade-off happens due to the fact that ETTh1/ETTh2 datasets are more challenging than ETTm1/ETTm2; thus, more memory is required to remain performing well in terms of MSE and still being parameter efficient compared to TAFAS.

\begin{figure*}[!ht]
    \centering
    \includegraphics[width=1.0\linewidth]{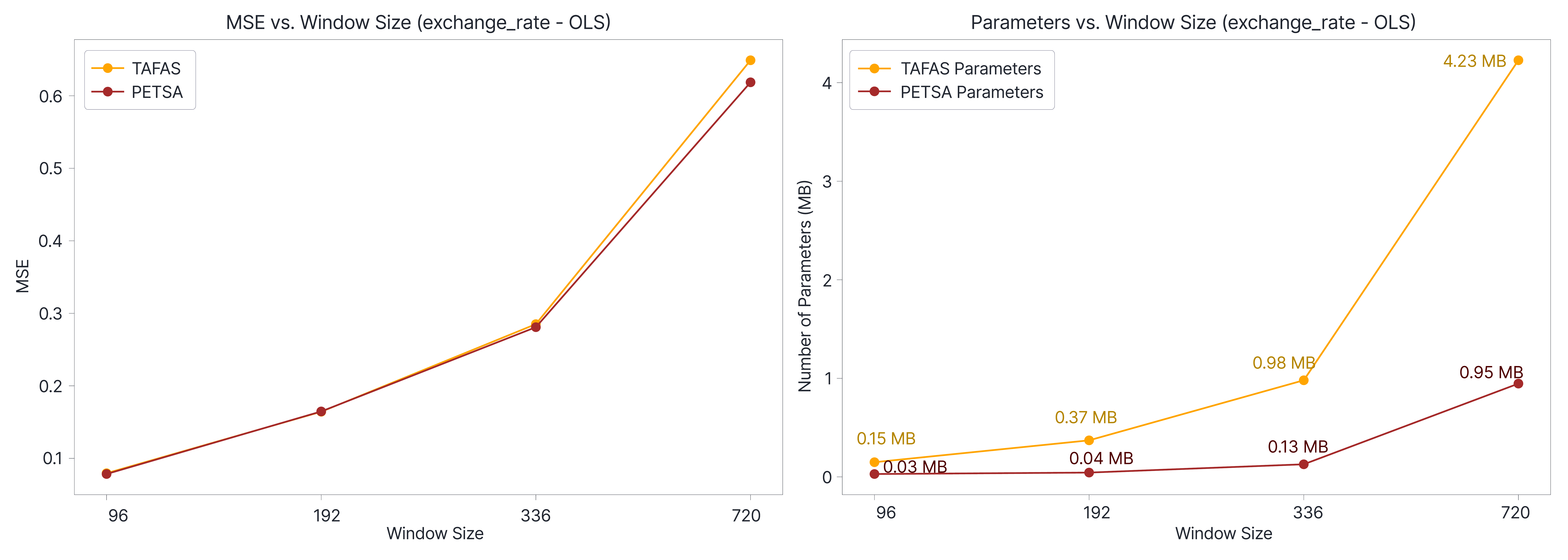}
    
    \caption{\textbf{Comparison of PETSA and TAFAS on Exchange Rate for OLS.} Left: MSE across different window sizes with TAFAS, and PETSA. Right: Number of trainable parameters used for adaptation. Memory usage is annotated in MB.}
    \label{fig:param_exchange_OLS}
\end{figure*}

\begin{figure*}[!ht]
    \centering
    \includegraphics[width=1.0\linewidth]{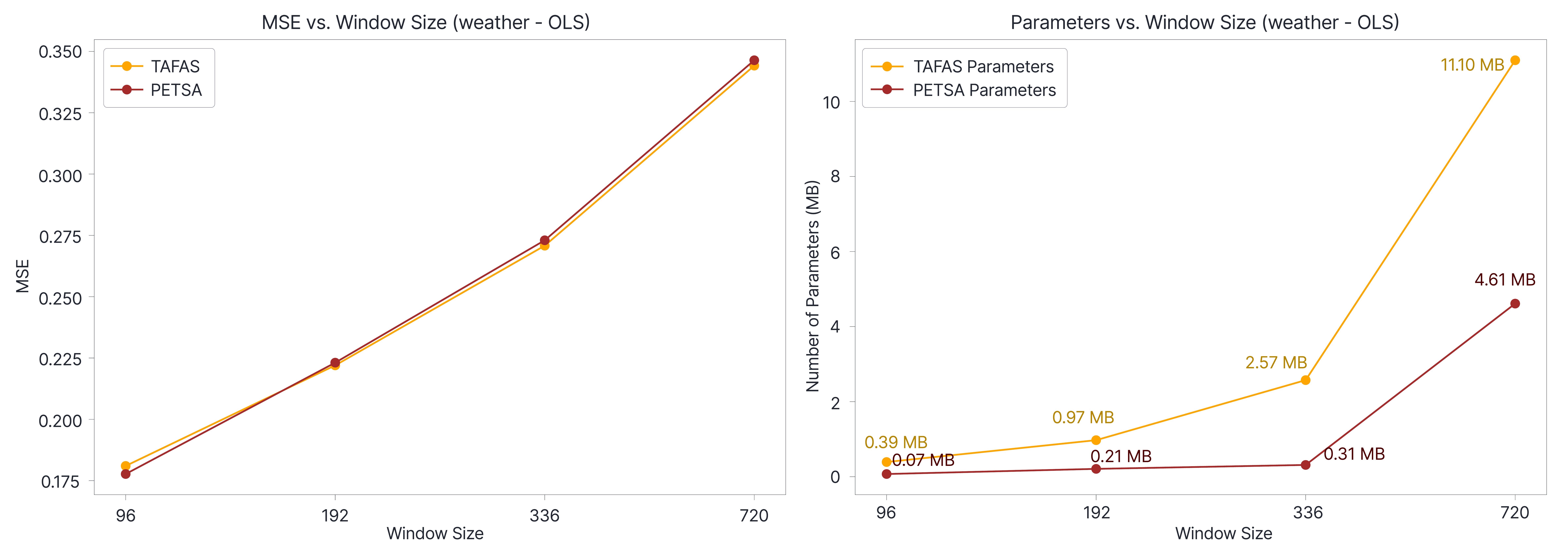}
    
    \caption{\textbf{Comparison of PETSA and TAFAS on Weather for OLS.} Left: MSE across different window sizes with TAFAS, and PETSA. Right: Number of trainable parameters used for adaptation. Memory usage is annotated in MB.}
    \label{fig:param_weather_OLS}
\end{figure*}

In Figure~\ref{fig:param_exchange_OLS}, PETSA and TAFAS show similar MSE trends across window sizes, with both methods degrading as the horizon increases. However, PETSA requires over 4× less memory than TAFAS at window size 720, making it a much more efficient alternative. Finally, in Figure~\ref{fig:param_weather_OLS}, despite similar performance, PETSA significantly reduces the adaptation cost, using less than half of the memory compared to TAFAS at window size 720.

\textbf{Ablation on Low-Rank (R) parameter}

\begin{figure*}[!ht]
    \centering
    \includegraphics[width=1.0\linewidth]{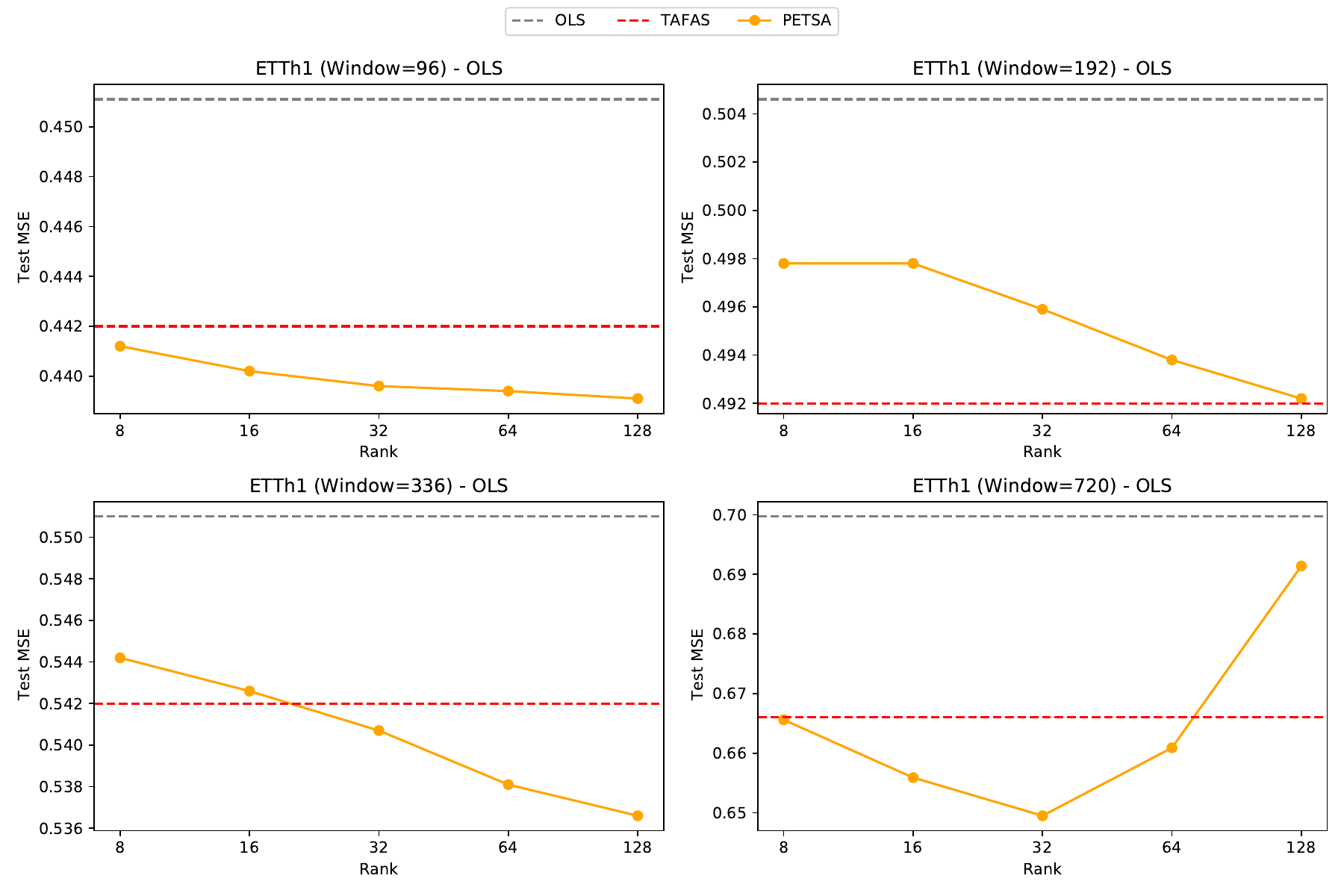}
    \caption{\textbf{Comparison of the original model, TAFAS, and PETSA on ETTh1 for OLS.} MSE across different ranks for windows 96, 196, 336, and 720.}
    \label{fig:low_rank_etth1_ols}
\end{figure*}

In this ablation, Figure~\ref{fig:low_rank_etth1_ols}, we conduct a study about the low-rank hyperparameter, which directly impacts the number of additional trainable parameters for the dynamic gating mechanism.

\textbf{Ablation on Dynamic Gating parameter}

\begin{figure*}[!ht]
    \centering
    \includegraphics[width=1.0\linewidth]{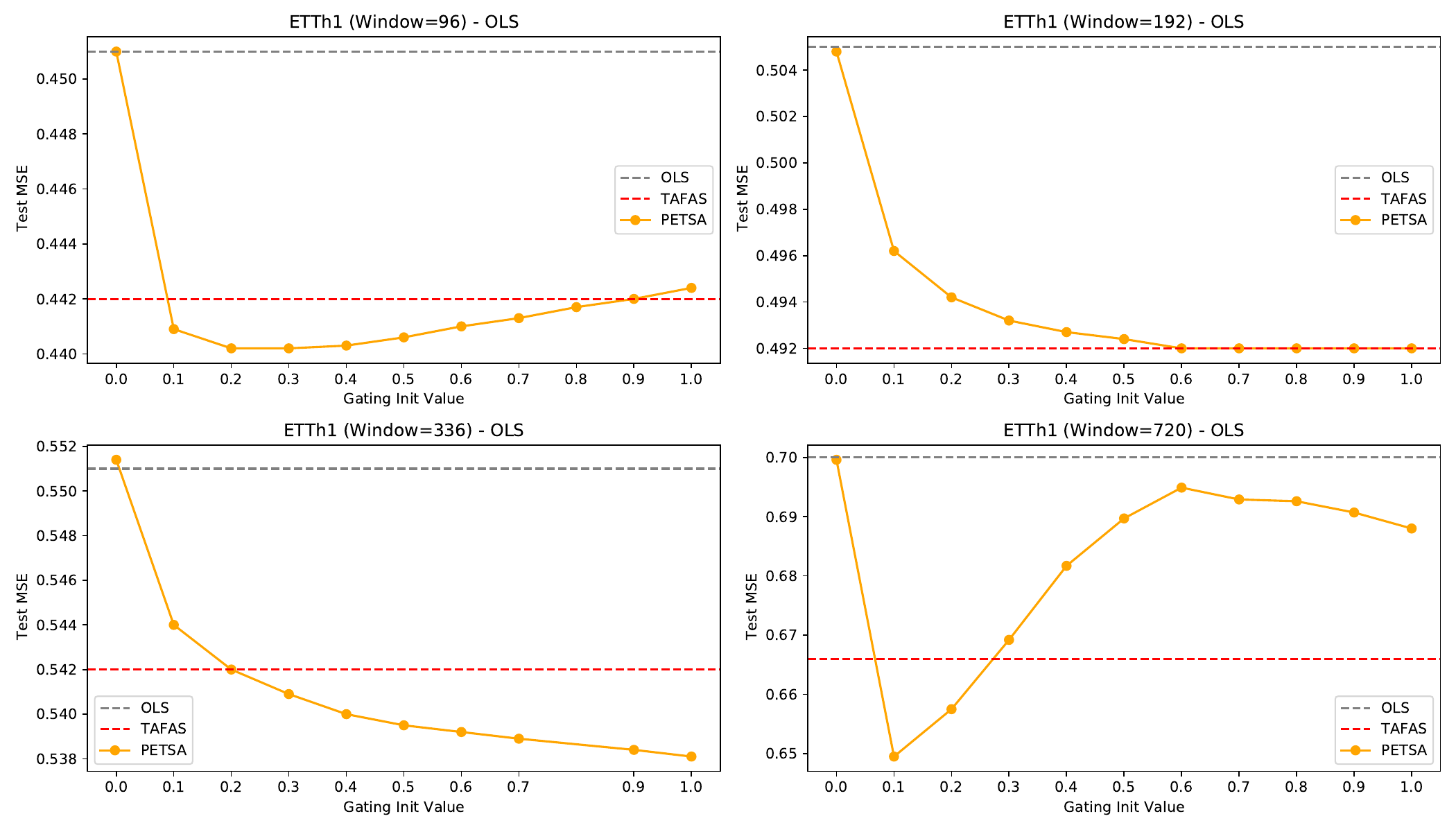}
    \caption{\textbf{Comparison of the original model, TAFAS, and PETSA on ETTh1 for OLS.} MSE across different gating initial values for windows 96, 196, 336, and 720.}
    \label{fig:ablation_gating_etth1_ols}
\end{figure*}

In this ablation, Figure~\ref{fig:ablation_gating_etth1_ols}, we study the initial value for the dynamic gating. This hyperparameter impacts the weights of the low-rank adaptation, providing a learnable way conditioned on the input to adjust its values; higher values make the weights positive due to the tanh; otherwise, lower values make the adapted weight negative, decreasing the value of the final calibrated input.

\textbf{Ablation on Loss Components}

\begin{figure*}[!ht]
    \centering
    \includegraphics[width=1.0\linewidth]{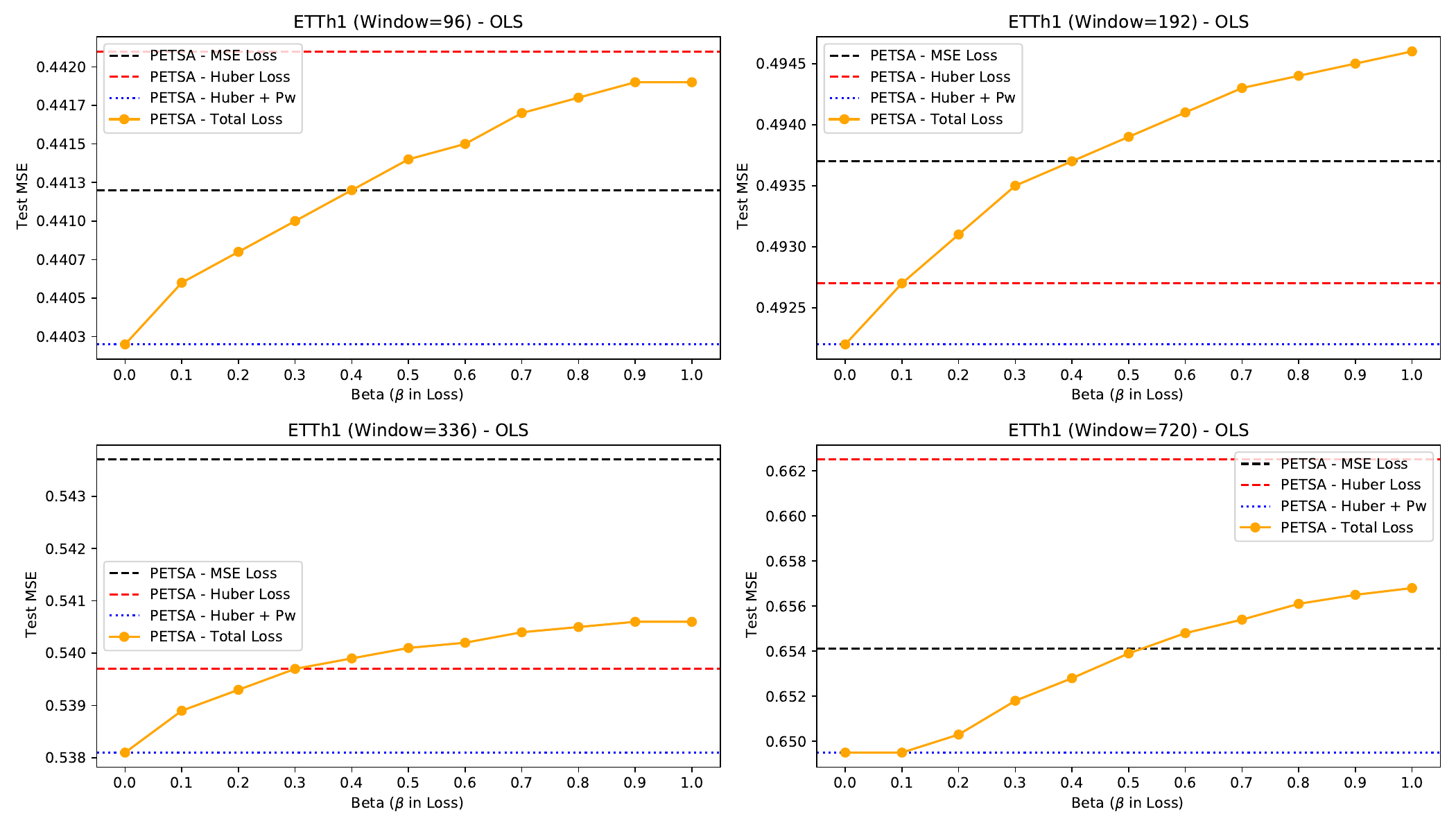}
    \caption{\textbf{Comparison of different loss terms in PETSA on ETTh1 for OLS.} MSE across different beta values for windows 96, 196, 336, and 720.}
    \label{fig:ablation_losses}
\end{figure*}

In this ablation, we study the impact of the loss components for PETSA during TTA. In Figure~\ref{fig:ablation_losses}, we see that the MSE loss is not sufficient for reaching the best performance values in terms of test MSE, similar to what occurs with only Huber loss. However, the total loss got the best results for ETTh1 OLS with $\beta$ equal to $0.0$, which means that the frequency component harmed the performance for this dataset, and $\beta=0.0$ means that only the Huber loss and structural patch components are being used. Depending on the model, the frequency loss helps the performance; for instance, the best performance for the FreTS model was when the $\beta$ was equal to $0.1$ (for the majority of the datasets and windows). For some datasets, a higher value can be the best result, so we recommend hyperparameter tuning for optimal performance.

